\pdfoutput=1
\documentclass[11pt]{article}

\usepackage{tabularx}
\usepackage{booktabs}
\usepackage{hyperref}
\usepackage{amsmath}
\usepackage{graphicx}
\usepackage{subfig}
\usepackage{CJKutf8}
\usepackage{algorithm}
\usepackage{algorithmic}

\usepackage{ACL2023}

\usepackage{times}
\usepackage{latexsym}

\usepackage[T1]{fontenc}
\usepackage[utf8]{inputenc}

\usepackage{microtype}

\usepackage{inconsolata}

\title{Sensitivity and Robustness of Large Language Models to Prompt Template in Japanese Text Classification Tasks}
\author{Chengguang Gan \And Tatsunori Mori \\
\AND
  Yokohama National University, Japan \\
  \texttt{gan-chengguan-pw@ynu.jp, tmori@ynu.ac.jp} \\}

\begin{document}
\maketitle
\begin{abstract}

Prompt engineering relevance research has seen a notable surge in recent years, primarily driven by advancements in pre-trained language models and large language models. However, a critical issue has been identified within this domain: the inadequate of sensitivity and robustness of these models towards Prompt Templates, particularly in lesser-studied languages such as Japanese. This paper explores this issue through a comprehensive evaluation of several representative Large Language Models (LLMs) and a widely-utilized pre-trained model(PLM). These models are scrutinized using a benchmark dataset in Japanese, with the aim to assess and analyze the performance of the current multilingual models in this context. Our experimental results reveal startling discrepancies. A simple modification in the sentence structure of the Prompt Template led to a drastic drop in the accuracy of GPT-4 from 49.21 to 25.44. This observation underscores the fact that even the highly performance GPT-4 model encounters significant stability issues when dealing with diverse Japanese prompt templates, rendering the consistency of the model's output results questionable. In light of these findings, we conclude by proposing potential research trajectories to further enhance the development and performance of Large Language Models in their current stage.

\end{abstract}

\section{Introduction}

With the advent of the transformer architecture\cite{vaswani2017attention}, language models have transitioned into the pre-training epoch. The concept of prompt-based learning emerged as a method for adapting downstream text classification tasks into fill-in-the-blank problems, thereby facilitating compatibility with the input requirements of these tasks\cite{schick2020exploiting}. This approach optimizes the alignment between the downstream task's input and the pre-training objective of BERT models\cite{devlin-etal-2019-bert}, specifically the Masked Language Model. Consequently, the model exhibits accelerated learning capabilities, even when exposed to a limited number of samples.

\begin{figure}[t]
\centering
\includegraphics[width=219 pt]{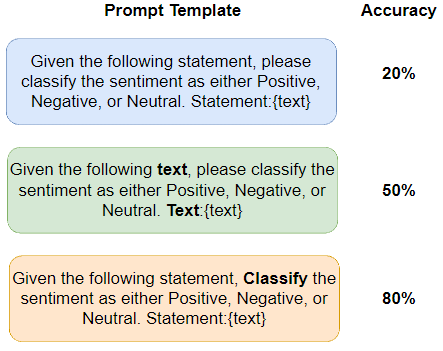}
\caption{\label{1figure1}The illustration depicts a comparison of accuracy among three distinct but semantically similar prompt template.}

\end{figure}

\par The manual methodology for designing prompts presents several limitations, as illustrated in Figure 1 using a sentiment classification task. Even though various prompts may appear semantically similar from a human perspective, they often yield considerably different results when implemented in the same model. Furthermore, the manual design process does not provide a definitive way to ascertain the optimal prompt template for a given task and model. Remarkably, a single prompt with ostensibly identical meaning can exhibit a wide array of performances. However, it is not feasible to experiment with all variations to identify the superior template. In subsequent research, Automated Prompt Template methods\cite{liu2023pre} like AutoPrompt\cite{shin-etal-2020-autoprompt} and Prefix-Tuning\cite{li-liang-2021-prefix} have been introduced. However, their applicability in the context of Large Language Models (LLMs) remains limited, hence, they will not be explored in this paper.
\par The introduction of ChatGPT\cite{ouyang2022training} has been met with considerable admiration due to its impressive performance. Nevertheless, questions persist concerning the model's sensitivity and robustness towards different prompt templates. Has ChatGPT effectively addressed these issues or at least mitigated them?
\par In addition to the extensive body of evaluation literature demonstrating the robustness of ChatGPT in major languages such as English and Chinese, there is a conspicuous lack of similar studies focused on less dominant languages. In this study, we focus on Japanese, one of the minor languages. Japanese is characterized by distinctive rules and grammatical constructs that set it apart from other languages.  These idiosyncrasies potentially complicate the construction of prompt templates, thereby posing unique challenges to the robustness and sensitivity of multilingual LLMs.  Consequently, a key question this research aims to answer is: how well can existing multilingual LLMs accommodate the prompt templates unique to diverse languages such as Japanese? 
\par To explore this question, we have undertaken a study which involves conducting experiments to assess the performance of existing LLMs for Japanese, a language that is less commonly modeled. This paper principally delivers the following key contributions:
1. A comprehensive evaluation of the robustness and sensitivity of LLMs is conducted concerning the influence of prompt templates on a Japanese benchmark dataset. The empirical findings disclose both the current advantages and shortcomings of LLMs.
2. The performance of LLMs is scrutinized through the application of an array of prompt templates, which indicates that the choice of specific words or phrases in Japanese prompt templates significantly impacts the LLMs' sensitivity. This insight, to a substantial degree, offers valuable guidance towards optimizing the model's training methodologies and corpus construction.
3. The experiments expose a substantial stability issue in LLMs when handling Japanese prompt templates of analogous meanings. This discovery subsequently introduces a novel avenue for enhancing the efficacy of future LLMs in Japanese.

\section{Related Work}

\par In this section, we elucidate upon existing studies that examine the performance of language models when faced with various prompts.

\par In the field of PLM. \cite{wangdomain} Examines the robustness of natural language models to domain shifts using prompt tuning and prefix tuning methods. Significant differences were found in domain robustness patterns between the two methods on T5 and GPT-2 models. The authors call for future research to explore the causes of these variations. In the realm of LLMs, prevailing research is predominantly centered on ChatGPT. One primary area of focus within this domain is the exploration of its application in translation tasks. Evaluates ChatGPT's ability to perform machine translation tasks\cite{jiao2023chatgpt}. The paper shows that ChatGPT performs competitively with commercial translation services on high-resource languages, but struggles with low-resource or distant languages. However, a pivot prompting strategy improves performance significantly. 
\par The comprehensive study by \cite{chen2023robust},  evaluates the robustness of GPT-3.5 on 21 datasets across 9 NLU tasks using TextFlint's text transformations. Findings show GPT-3.5 outperforms fine-tuned models in some tasks but experiences significant performance drops in others.  Specific challenges include robustness instability, prompt sensitivity, and number sensitivity.  The study also highlights variations in performance between different GPT models and the need for consistency in task labels and label types.  These insights are crucial for understanding limitations and guiding future research to improve AI model performance and generalization abilities. However, limits its scope to the examination of mainstream Chinese and English languages. It does not delve into a comprehensive evaluation and analysis of other less predominant languages.
\par There is also some work using LLMs, with Japanese as the research objectives. For example, testing the performance of GPT-4\cite{openai2023gpt4} and ChatGPT in Japanese Medical Licensing Examinations\cite{kasai2023evaluating}. Furthermore, using the GPT-3 model and multilingual prompt method for Japanese natural language tasks, above baseline results were achieved\cite{songlarge}. 
\par While previous experiments have provided valuable insights, they have not delved into the sensitivity and robustness of LLMs pertaining to the specific linguistic nuances of the Japanese language. Our primary objective is to \textbf{investigate the impact of subtle variations of prompt template on }

\begin{table*}[!t]
\centering
\setlength{\tabcolsep}{9pt}

\begin{tabular}{lcc}
\toprule[2pt]
  \textbf{Dataset} & \textbf{Category} & \textbf{Total number of categories}\\
\midrule
\textbf{MARC-ja} & positive,negative & 2\\
\textbf{JNLI} & contradiction,neutral,entailment & 3 \\ 
 % & ,entailment &  \\
\textbf{JSTS} & 0,1,2,3,4,5 & 6  \\
\bottomrule[2pt]
\end{tabular}
\caption{\label{table1}
Details of the Datasets (Total refers to the number of categories)}
\end{table*}

\textbf{model performance, especially when the meanings of the prompt templates bear a close resemblance}.

\section{Experiment Setup}

This section commences with an introduction to the general parameters of the experiment. Subsequently, we delve into the three pivotal aspects of the experiment specifically the dataset, the model, and the design of the prompt template in a more comprehensive manner.
\par The experiments are initially established as zero-shot. This is primarily due to the fact that, in most real-world implementations of LLMs, the responses are provided directly to the users' inquiries, without any domain-specific fine-tuning. Consequently, our experiment will exclude the utilization of few-shot or In-context Learning methodologies\cite{dong2023survey}. The model will not be provided with a set of examples to learn from. Adopting a zero-shot approach also circumvents the issue of bias that could arise when selecting samples for a few-shot experiment.
\par In regard to the model's output settings, the maximum number of new tokens is limited to 10. Additional parameters are not deliberately defined at this stage, as our aim is to control the variables of the experiment as rigorously as possible. This is done to safeguard the fairness and integrity of the overall comparative study. Furthermore, due to constraints in time and resources, we have selected a test set size of 1000. The random seed has been set at 42, which is arbitrarily selected from the dataset.

\subsection{Dataset}

In our dataset selection, we employed the Japanese General Language Understanding Evaluation (JGLUE)\cite{kurihara-etal-2022-jglue} benchmark, which encompasses five distinct datasets: text classification, sentence pair classification, and QA tasks.  We focused on three specific datasets predominantly geared towards text classification and sentence pair classification tasks, given the fundamental importance of sentence classification within Natural Language Processing (NLP) as an effective gauge of a model's text comprehension capabilities.

\par A secondary rationale behind selecting these three text classification tasks pertains to their escalating difficulty level in terms of categories, as delineated in Table \ref{table1}.  The initial task, MARC-ja, revolves around binary sentiment classification.  This is followed by Japanese Natural Language Inference (JNLI), a tri-category sentence pair classification task, and concludes with the Japanese Semantic Textual Similarity (JSTS), initially a 26-category task ranging from 0-5 at 0.2 intervals.  For the sake of maintaining a linear progression in difficulty, we modified JSTS into a 6-category task with a 0-5 score range at 1 point intervals.

\begin{table}[!h]
\centering
\setlength{\tabcolsep}{8pt}

\begin{tabular}{llc}
\toprule[2pt]
  \textbf{Model} & \textbf{Parameters} & \textbf{Order}\\
\midrule
\textbf{T5-base} & 220 Million & $10^8$ \\
\textbf{LLaMA-7B} & 7 Billion & $10^9$ \\ 
\textbf{LLaMA-7B-LoRA} & 7 Billion & $10^9$  \\
\textbf{LLaMA-13B} & 13 Billion &  $10^{10}$ \\
\textbf{GPT-3.5-Turbo} & — &  — \\
\textbf{GPT-4} & — &  — \\
\bottomrule[2pt]
\end{tabular}
\caption{\label{table2}
Detail of Models.}
\end{table}

\subsection{Model}

Table \ref{table2} presents several mainstream models selected based on their exponential size. To benchmark against PLM era models, we chose T5-base-japanese\cite{raffel2020exploring}\footnote{\raggedright\url{https://huggingface.co/sonoisa/t5-base-japanese}}, representative of the hundred-million-level. At the billion and ten-billion levels, we selected the open-source LLaMA\cite{touvron2023llama} model. Given that LLaMA is primarily trained on English corpora, its support for Japanese might be limited. For comparison, we included LLaMA-7B-LoRA\footnote{\raggedright\url{https://huggingface.co/KBlueLeaf/guanaco-7b-leh-v2}}, a model

\begin{figure*}[!ht]
\centering
\includegraphics[width=450 pt]{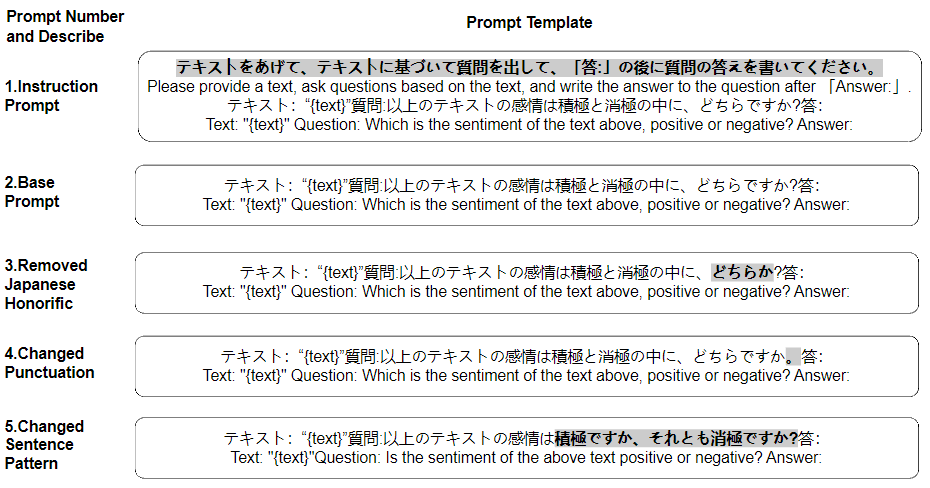}
\caption{\label{2figure2}The illustration of designed prompt template.}

\end{figure*}

enhanced with training on Chinese and Japanese corpora using the LoRA adapter. This selection aids in assessing whether simply training on Chinese and Japanese corpora improves Japanese task performance. Our approach ensured test models with exponentially increasing parameters, attempting to corroborate that model performance also increases linearly.

\subsection{Prompt Template Design}

In this section, we delve into the process of crafting distinct prompt templates to evaluate the performance of the language model concerning Japanese linguistic features.  As demonstrated in Figure 2, we utilize the initial sentiment classification dataset as an example, with templates numbered from 1 to 5, each representing a different design strategy.
\par Template No. 1 was conceived using the concept of Instruction Learning\cite{wei2021finetuned}.  Here, explicit instructions are appended to the basic questions to stimulate the language model's comprehension.  However, this varies from the original paper which presented multiple choices(e.g., "A","B","C","D" ) for classification, requiring the model to output the same for prediction.  To better align with real-world application scenarios, we adopted a narrative style of questioning where the options are seamlessly integrated into the question sentences.
\par Template No. 2 constitutes the fundamental prompt template, albeit devoid of the instructive sentence.  In contrast, Template No. 3 retains the basic prompt structure but omits the distinctive Japanese honorific word "\begin{CJK}{UTF8}{min} です\end{CJK}";  a deletion that does not alter the overall sentence meaning.
\par Template No. 4 presents a minor change, with the question mark "?" replaced by the period "\begin{CJK}{UTF8}{min} 。\end{CJK}".  This modification reflects the characteristic of Japanese interrogative sentences which can end with either a question mark or a period.
\par Lastly, Template No. 5 maintains the original sentence meaning but modifies the syntax of the sentence's latter half.  Our objective is to determine if these nuanced differences exert any significant impact on the model's ultimate performance.

\section{Evaluation}

\begin{table*}[ht]
\centering

\begin{tabularx}{1\textwidth}{@{}l*{6}{>{\centering\arraybackslash}X}@{}}
\toprule[2pt]
 & & & MARC-ja & &\\
Prompt Template & 1 & 2 & 3 & 4 & 5 & SD \\
\midrule
T5-base & 3.00(34.74) & 49.80(12.06) & 44.40(6.66) & 47.60(9.86) & 43.90(6.16) & 17.50\\
LLaMA-7B & 54.30(16.82) & 80.80(9.68) & 80.10(8.98) & 81.10(9.98) & 59.30(11.82) & 11.80\\
LLaMA-7B-LoRA & 82.40(0.38) & 84.60(2.58) & 83.60(1.58) & 83.90(1.88) & 75.60(6.42) & 3.29\\
LLaMA-13B & 83.00(13.02) & 65.00(4.98) & 73.40(3.42) & 66.30(3.68) & 62.20(7.78) & 7.48\\
GPT-3.5-Turbo & 71.10(5.52) & 77.20(0.58) & 77.40(0.78) & 76.80(0.18) & 80.60(3.98) & 3.08\\
GPT-4 & \textbf{88.70}(0.26) & \textbf{88.70}(0.26) & \textbf{86.20}(2.24) & \textbf{88.50}(0.06) & \textbf{90.10}(1.66) & 1.26\\
\bottomrule[2pt]
\end{tabularx}
\vspace{1em}

\begin{tabularx}{1\textwidth}{@{}l*{6}{>{\centering\arraybackslash}X}@{}}
\toprule[2pt]
 & & & JNLI & &\\
Prompt Template & 1 & 2 & 3 & 4 & 5 & SD \\
\midrule
T5-base & 1.00(0.70) & 0.30(0.00) & 0.00(0.30) & 0.20(0.10) & 0.00(0.30) & 0.37\\
LLaMA-7B & 14.40(7.68) & 18.70(3.38) & 37.10(15.02) & 17.20(4.88) & 23.00(0.92) & 8.01\\
LLaMA-7B-LoRA & 15.10(8.98) & 15.70(8.38) & 18.30(5.78) & 17.10(6.98) & \textbf{54.20}(30.12) & 15.10\\
LLaMA-13B & 19.60(6.14) & 11.10(2.36) & 13.00(0.46) & 10.80(2.66) & 12.80(0.66) & 3.19\\
GPT-3.5-Turbo & 31.60(12.6) & \textbf{49.70}(5.5) & \textbf{43.10}(1.1) & \textbf{48.70}(4.5) & 47.90(3.70) & 6.70\\
GPT-4 & \textbf{40.39}(6.75) & 25.54(8.10) & 27.62(6.02) & 25.44(8.2) & 49.21(15.57) & 9.56\\
\bottomrule[2pt]
\end{tabularx}
\vspace{1em}

\begin{tabularx}{1\textwidth}{@{}l*{6}{>{\centering\arraybackslash}X}@{}}
\toprule[2pt]
 & & & JSTS & &\\
Prompt Template & 1 & 2 & 3 & 4 & 5 & SD \\
\midrule
T5-base & 0.00(0.88) & 1.60(0.72) & 1.70(0.82) & 0.40(0.48) & 0.70(0.18) & 0.67\\
LLaMA-7B & 11.30(0.86) & 11.90(0.26) & 11.80(0.36) & 12.30(0.14) & 13.50(1.34) & 0.74\\
LLaMA-7B-LoRA & 29.30(10.62) & 14.30(4.38) & 10.10(8.58) & 15.80(2.88) & 23.90(5.22)  & 6.94\\
LLaMA-13B & 9.70(1.82) & 12.00(0.48) & 11.90(0.38) & 11.90(0.38) & 12.10(0.58) & 0.91\\
GPT-3.5-Turbo & 34.50(3.18) & 39.90(2.22) & 38.30(0.62) & 39.70(2.02) & 36.00(1.68) & 
 2.11\\
GPT-4 & \textbf{49.20}(2.21) & \textbf{48.37}(1.38) & \textbf{50.70}(3.71) & \textbf{49.55}(2.56) & \textbf{37.11}(9) & 4.99\\
\bottomrule[2pt]
\end{tabularx}

\caption{\label{table5}
The accuracy of 0-shot experiment with different prompt templates for MARC-ja, JNLI, and JSTS. The value in "()" is the absolute deviation. SD is denote Standard Deviation.
}
\end{table*}

We employed a text classification methodology utilizing a question and answer format. Consequently, the generated output from the model could potentially be a complete sentence, such as "It's positive," which incorporates the designated label word, in this case, "positive." To extract this classified label word from the generated response text, we devised a straightforward algorithm. For the assessment of model performance, we selected accuracy as the metric to facilitate evaluation.
The pseudo-code delineating the process of extracting label words from the generated text can be located in Appendix \ref{sec:algorithm}.

\par Moreover, to highlight the variability of accuracy across different models for varying prompt templates, we conducted a statistical analysis. Assuming the model's accuracy on prompt templates No.1-5 as $(x_1, x_2, x_3, x_4, x_5)$, we calculated their mean value, denoted as $M$. Subsequently, we determined the absolute deviation score by calculating the absolute value of the difference between each $x$ value and $M$. Through the computation of the absolute deviation score, one can effectively illustrate the extent to which each accuracy deviates from the average.

\begin{equation}
M = \frac{x_1 + x_2 + x_3 + x_4 + x_5}{5}
\end{equation}

\begin{align}
\text{Absolute Deviation Score} &= |x_i - M| \nonumber \\
&\text{for } i = 1,2,3,4,5
\end{align}

\par In addition to this, we computed the standard deviation of a set of accuracies for each model, thereby providing a statistical depiction of the accuracy fluctuations across different prompt templates for each model.

\section{Results}

In this section, we present a segmented analysis of the experimental results, divided according to the specific datasets used. Each segment will be examined individually. Notably, an in-depth exploration of the results concerning the T5-base will be provided in Section \ref{6.2} as a representative example.
\par Table 1 summarizes the outcomes of all conducted experiments. A review of the overall results reveals a clear correlation between the high and low accuracy rates across data sets and the performance capabilities of each model. Nonetheless, a number of anomalous findings emerged. These unexpected results offer valuable insights, potentially highlighting areas of concern within the current LLMs.

\subsection{MARC-ja}

In the basic sentiment binary classification task using the MARC-ja dataset, the GPT-4 model demonstrated the lowest standard deviation at 1.26. This implies that GPT-4's performance across different prompt templates was relatively consistent when compared to other models. However, there were noticeable deviations in the performance of GPT-4, with an absolute deviation score of 2.24 for Prompt Template 3 and 1.66 for Prompt Template 5. It was observed that the omission of an honorific from Prompt Template 3 resulted in a decrease in accuracy to 86.20. Conversely, a modification in the sentence structure of Prompt Template 5 led to an increased accuracy of 90.10. This data suggests that the stability of GPT-4 when processing various prompt templates remains an area for improvement.
\par Secondly, the LLaMA-7B-LoRA model, trained using the Sino-Japanese corpus, demonstrated a significant reduction in standard deviation to 3.29 approximately half that of LLaMA-13B. Furthermore, its accuracy outstripped that of LLaMA-7B by a remarkable 28.1 points on the instruct Prompt Template 1. This underscores the model's capability to enhance the comprehension of minor languages like Japanese, given expanded training on Chinese and Japanese corpus. Yet, after the change in Prompt Template 5, LLaMA-7B-LoRA experienced notable volatility, with accuracy tumbling to 75.60. Despite this, the LLaMA-7B-LoRA model's overall performance surpassed that of GPT-3.5-Turbo on the MARC dataset.
\par 
The subsequent findings for LLaMA-13B illustrate a compelling phenomenon. It appears that nearly doubling the size of the model precipitates a substantial enhancement in the accuracy of instruct Prompt Template 1, which escalates dramatically to 83. This finding suggests a proportional correlation between the model's size and its ability to comprehend the problem more proficiently. The improvement in performance is indicative of the critical role instruct learning plays, thus reaffirming its influence in the LLMs learning process.

\subsection{JNLI}

Upon increasing the task difficulty to triple classification in the JNLI dataset, a significant performance drop is observed across all models. The LLaMA-13B and T5-base models exhibit small standard deviations, which could be attributed to their low accuracy. The performance degradation for the remaining four models is even more pronounced. Notably, the standard deviations for GPT-3.5-Turbo and GPT-4 models escalated to 6.7 and 9.56, respectively. Moreover, the GPT-3.5-Turbo model managed to surpass the GPT-4 model by approximately twofold in the accuracies of Prompt Templates 2-4. Nevertheless, despite these fluctuations, both the GPT-3.5-Turbo and GPT-4 models consistently demonstrated superior performance when compared to the LLaMA model.

\subsection{JSTS}

In summary, there is a noticeable reduction in the standard deviation when dealing with the complex six-category JSTS dataset.   The ranking of model accuracy, both at its highest and lowest, appears to be aligned with the inherent strengths and weaknesses of the respective models.   It is notable that the GPT-3.5-Turbo model demonstrated a significant decline in performance, with the accuracy dropping to 34.5 when utilizing construct Prompt Template 1.   Similarly, the GPT-4 model experienced a substantial decrease, with accuracy falling to 37.11 upon employing the modified Prompt Template 2. Nevertheless, the complexity of the JSTS dataset, particularly with the sentence type alterations in the modified Prompt Template 5, exceeded those in the previous four prompts template. This increased complexity led to the observed decline in accuracy.

\section{Analysis}

In this section, we present a comprehensive analysis and interpretation of the experimental results, aiming to enhance their intuitiveness.  The analysis of these results will be conducted in two parts.  The first subsection \ref{6.1} involves analysis through the use of line graphs and box-plots to demonstrate the data's distribution and trends.  In the second subsection \ref{6.2}, we integrate the actual textual content of the generated results into our analysis, providing a more contextual understanding of the findings.

\subsection{Analysis with Plot}\label{6.1}

\begin{figure*}[ht]
    \centering
    \subfloat[Accuracy for MARC-ja]{%
        \includegraphics[width=0.31\textwidth]{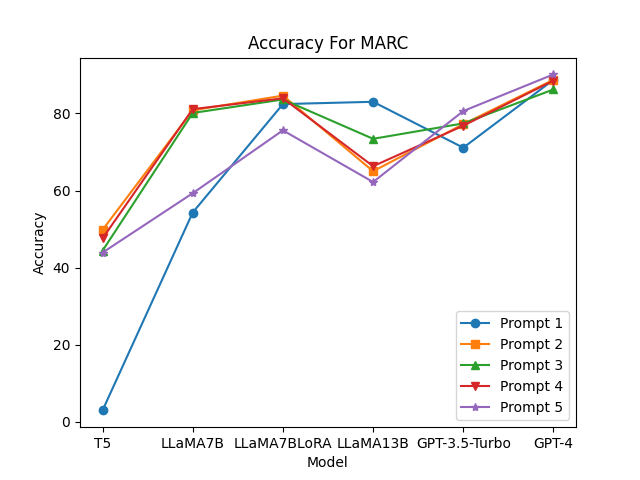}%
        \label{fig3:a}%
    }\hfill
    \subfloat[Accuracy for JNLI]{%
        \includegraphics[width=0.31\textwidth]{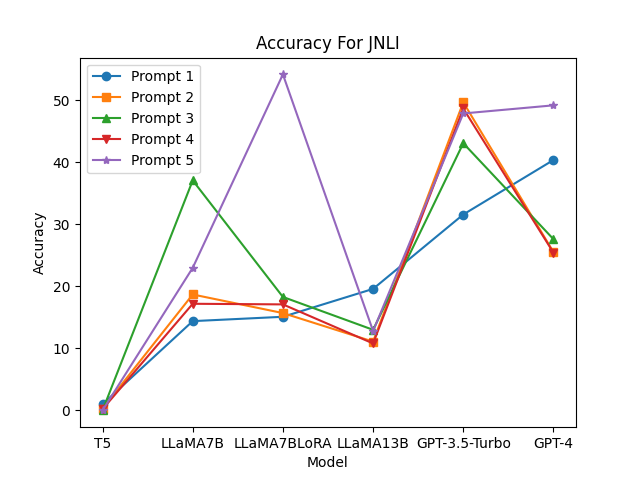}%
        \label{fig3:b}%
    }\hfill
    \subfloat[Accuracy for JSTS]{%
        \includegraphics[width=0.31\textwidth]{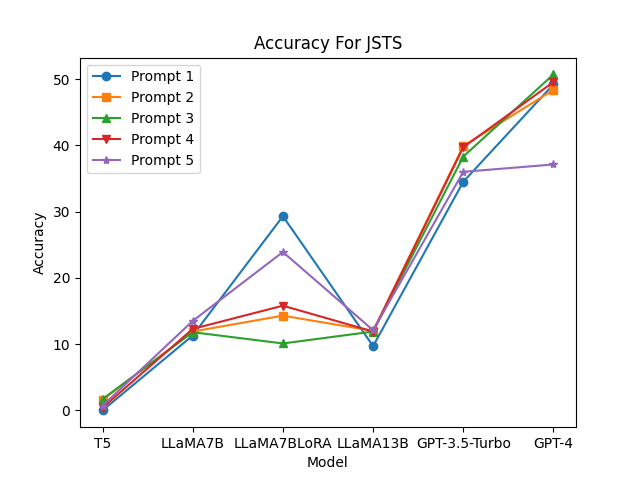}%
        \label{fig3:c}%
    }
    \caption{Compare the accuracies of the same prompt template on different models.}
    \label{fig3:line}
\end{figure*}

\begin{figure*}[ht]
    \centering
    \subfloat[Accuracy for MARC-ja]{%
        \includegraphics[width=0.31\textwidth]{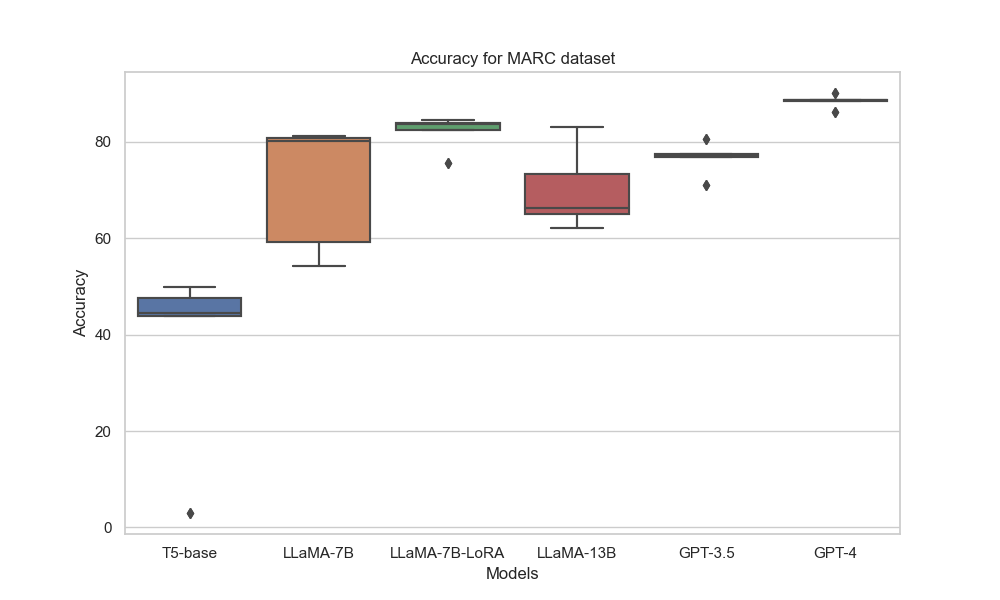}%
        \label{fig4:a}%
    }\hfill
    \subfloat[Accuracy for JNLI]{%
        \includegraphics[width=0.31\textwidth]{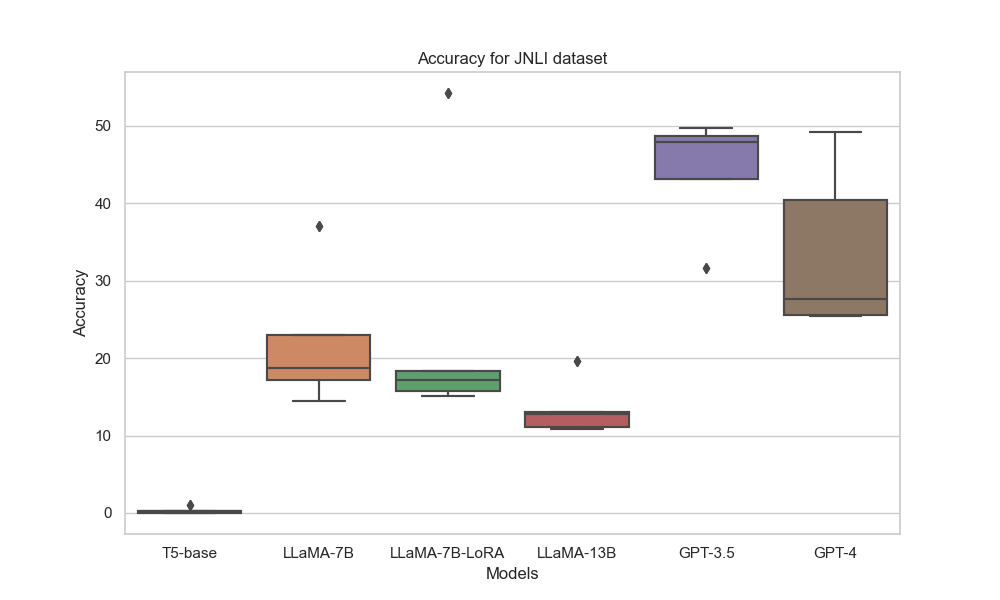}%
        \label{fig4:b}%
    }\hfill
    \subfloat[Accuracy for JSTS]{%
        \includegraphics[width=0.31\textwidth]{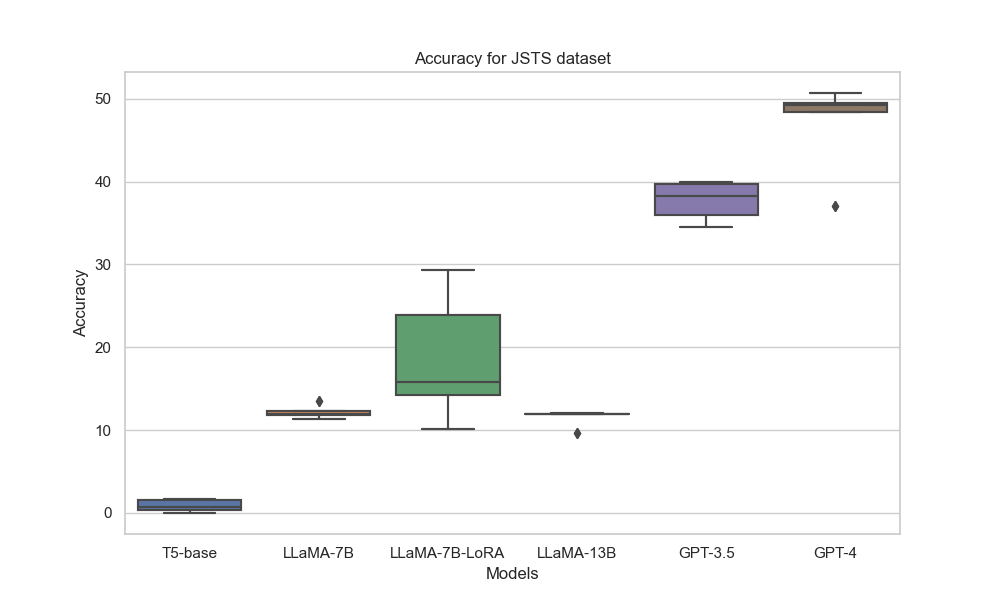}%
        \label{fig4:c}%
    }
    \caption{Compare the fluctuations of different models on the five prompt templates.}
    \label{fig4:box}
\end{figure*}

Figure \ref{fig3:line} presents three line graphs, each corresponding to a different difficulty dataset.  The horizontal axis denotes various models, while the vertical axis indicates accuracy.  Ideally, there should be a broadly increasing trend aligned with model performance strength.  Yet, as demonstrated in Figure \ref{fig3:a}, Prompt Template 1 exhibits robust performance with both the LLaMA series and GPT-4 models on the MARC-ja dataset, despite a decrease in accuracy with the GPT-3.5-Turbo model.  Upon examining the other four prompt templates of GPT-3.5-Turbo, it becomes evident that adding an instructive sentence to Prompt Template 1 results in a performance degradation.

\par Figure \ref{fig3:b} demonstrates a significant increase in the model's sensitivity towards the prompt template, particularly as the classification task complexity escalates. Notably, the model's robustness seems to exhibit considerable variability. Nevertheless, it is essential to highlight that of all the prompt templates, Prompt Template 1 remains the most stable. The accuracy fold displays a steady upward trend, which underscores the efficacy of instruction learning for LLMs. These models exhibit a superior capacity for understanding long text problems in comparison to the T5-base model. However, the scenario is different for GPT-3.5-Turbo and GPT-4. There are prominent fluctuations across multiple prompt templates, suggesting that even minor differences can lead to significant variations in outcomes, especially for slightly more complex tasks. This scenario is particularly prevalent when ChatGPT products are in use; user's habitual usage of a specific prompt template may yield effective results on one model, but switching to a different model may result in a subpar performance. This issue warrants immediate attention, as it presents a pressing challenge for the model's overall utility and effectiveness.

\par As the complexity of the task escalates incrementally, the GPT series model consistently outperforms the LLaMA series models, as clearly depicted in Figure \ref{fig3:c}. Furthermore, the GPT series model demonstrates superior stability when processing diverse prompt templates, with negligible fluctuation observed.
\par To more effectively illustrate the degree of fluctuation exhibited by various models when processing different prompt templates, we utilized box-plots. These diagrams offer a visual representation of model fluctuation for each prompt template, with the area of the box indicating the extent of the variation. As demonstrated in Figure \ref{fig4:box}, certain data points are observed outside the boundaries of the box-plot.  This occurrence can be attributed to the nature of the box-plot representation, which intentionally excludes certain outliers, thereby providing a more accurate depiction of the underlying data's volatility. 
The box-plot features additional extended lines, which represent the maximum and minimum values once outliers have been excluded.
\par As depicted in Figure \ref{fig4:a}, the LLaMA-7B-LoRA model achieves performance parity with the GPT series, even exceeding the accuracy of GPT-3.5-Turbo. This demonstrates the efficacy of LLaMA-7B-LoRA, particularly after training with Chinese-Japanese language enhancement, in significantly improving performance for the less commonly used Japanese language. This improvement is noteworthy, as it is achieved without the necessity for a vast quantity of parameters, unlike ChatGPT.
\par In contrast, Figure \ref{fig4:b} reveals that despite the significant performance enhancement of the GPT series when compared to other models, there is a marked increase in the area of the box-plot. This suggests potential robustness issues when these models handle different prompt templates.
\par Lastly, Figure \ref{fig4:c} shows that with the exception of the LLaMA-7B-LoRA model, which displays significant stability fluctuations, the box-plot area of all other models remains relatively small, indicating lesser degrees of fluctuation.

\begin{figure}[!h]
\centering
\includegraphics[width=219 pt]{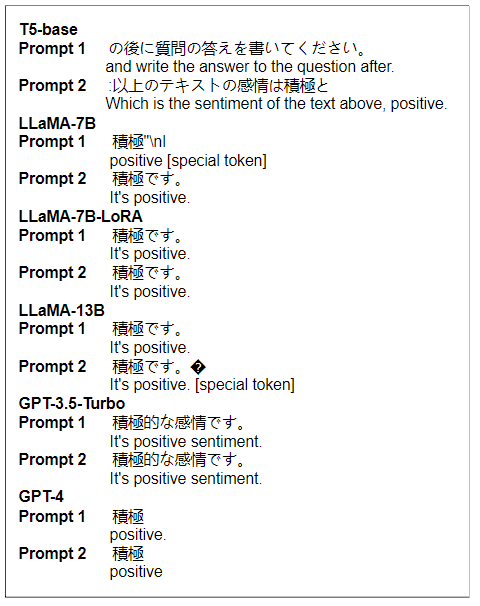}
\caption{\label{6figure6}The generated text example of MARC-ja dataset.}

\end{figure}

\subsection{Analysis with Generated Text}\label{6.2}

Figure \ref{6figure6} illustrates the specific results generated from Prompt Templates 1 and 2 applied to the MARC-ja dataset. A notable observation is that the T5 model fails to comprehend the overarching meaning of the question, hence being unable to produce a relevant answer. Instead, it merely replicates the sentence following "Answer:" from the input text as the output. This outcome is likely attributed to the model's limited size, which results in the generation of tokens based on maximum probability, rather than contextual understanding.
\par In contrast, LLMs, such as LLaMA-7B and LLaMA-13B, successfully interpret the questions and respond appropriately. Despite the lack of native support for Japanese within these models, they manage to generate meaningful responses. However, they occasionally produce unique tokens, which is likely a byproduct of the linguistic limitation. Furthermore, the incorporation of the Japanese corpus in the LLaMA-7B-LoRA model training significantly enhances its effectiveness in generating Japanese text.
\par Finally, it is worth highlighting the difference in language style between GPT-3.5-Turbo and GPT-4. GPT-3.5-Turbo's output bears a closer resemblance to conversational language, while GPT-4 adeptly generates responses in a format aligned with the specifications given in the question.

\subsection{Limitations}

 This work was tested only for Japanese. It was not tested for other minor languages. Furthermore, due to limitations in time and resources, this experiment could not be conducted utilizing the full extent of the Japanese text classification dataset. Secondly, the experiments were conducted only for the basic text classification task. And no other natural language processing tasks were tested. In terms of model selection, we opted for the most recent and representative LLMs.  However, we acknowledge that this choice may not yield extensive coverage.  Despite these limitations, we posit that our approach helped control extraneous variables, thus presenting a more realistic, controlled experimentation environment that closely reflects real-world conditions.

\section{Conclusion}

This study endeavors to evaluate the sensitivity and resilience of existing Large Language Models (LLMs) in response to varying prompt templates, utilizing a basic Japanese benchmark dataset as a test case. The experimental findings indicate four key insights: 1) Current LLMs exhibit a degree of instability, with even slight variations in the prompt template potentially leading to substantial fluctuations in model output; 2) Uniform applicability of a single prompt template across all LLMs is not guaranteed; 3) Instruction-based learning proves to be more effective for some LLMs(e.g., LLaMA-13B) when compared to the use of a common prompt template; 4) LLMs require additional training to bolster their robustness when dealing with the Japanese language.
\par These findings pave the way for a significant research direction for LLM developers and researchers, namely, \textbf{enhancing the resilience of LLMs to diverse prompt templates when dealing with less commonly studied languages}. We assert that addressing this issue is vital for the future progression of LLMs and represents a challenge that necessitates resolution.

\bibliography{custom}
\bibliographystyle{acl_natbib}

\appendix

\begin{figure*}[!ht]
\centering
\includegraphics[width=450 pt]{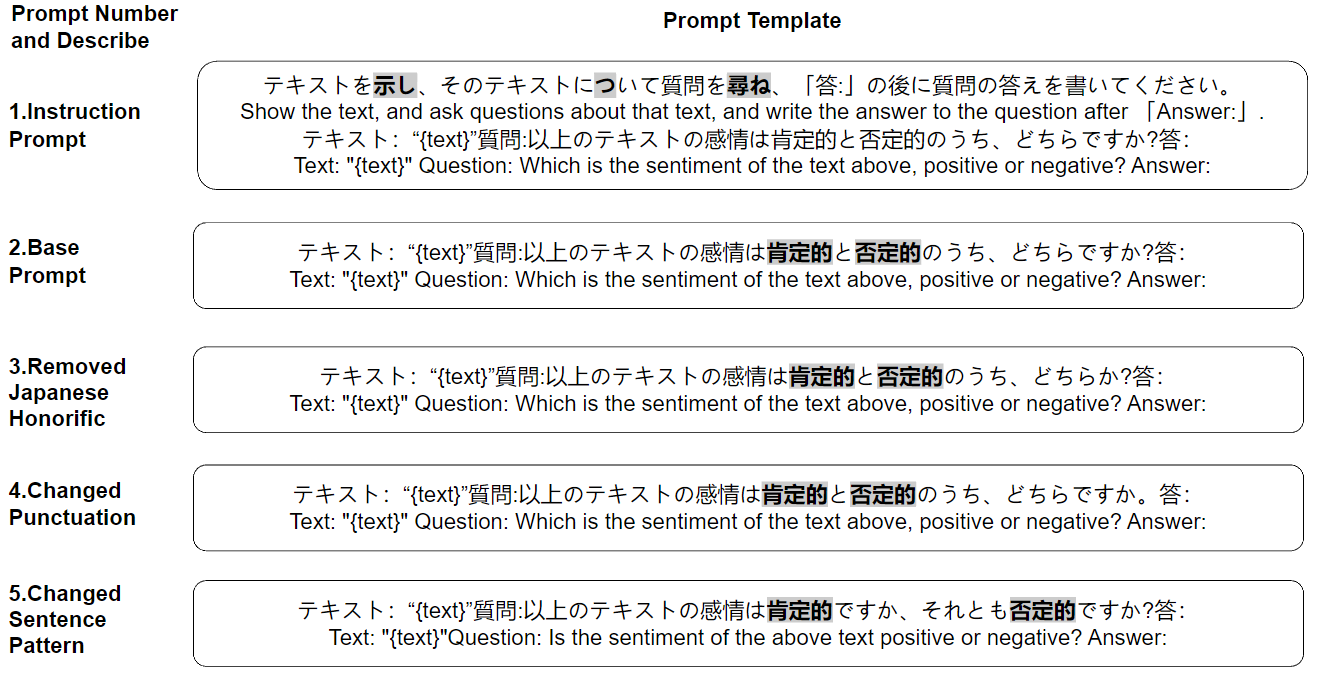}
\caption{\label{7appendixfigure1}The example of prompt template designed from the perspective of native Japanese speakers.}

\end{figure*}

\section{Prompt Template in Native Speaker}
\label{sec:native}

As shown in Figure \ref{7appendixfigure1}, it is a more native Japanese question sentence.
But in this experiment, we did not use this sentence as prompt template.
First of all, Japanese and Chinese belong to the same Kanji circle. If we could replace the positive and negative words with Chinese words that also exist in the Chinese language. The model might be able to gain cross-linguistic knowledge from the richer Chinese corpus to enhance the understanding on Japanese. So we replaced several kanji words with kanji words common to both Chinese and Japanese in the native language representation here.

\section{Prompt Template of JNLI and JSTS}
\label{sec:prompttemplate}

As shown in Figure \ref{7appendixfigure1}, the prompt template we designed for the JNLI dataset. The design idea of the five different prompt templates for the three datasets is the same. A category of words is added to the previous MARC-ja dataset.

\par
\begin{figure*}[!t]
\centering
\includegraphics[width=450 pt]{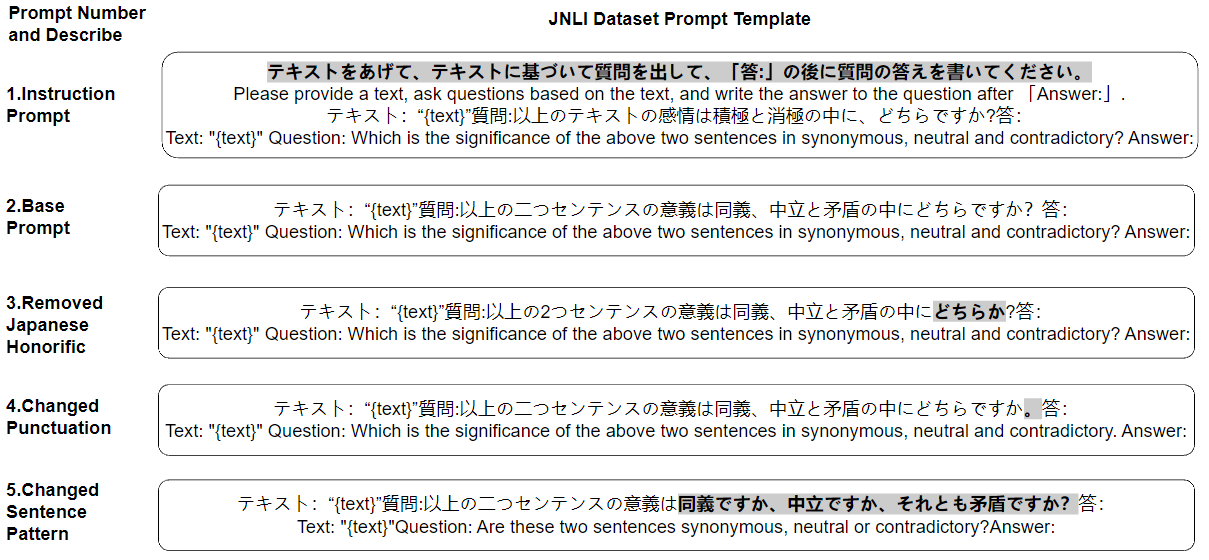}
\caption{\label{8appendixfigure2}The designed prompt template of JNLI dataset.}

\end{figure*}

 In contrast, Figure \ref{8appendixfigure2} demonstrates a slight deviation from this pattern due to the inherent structure of the JSTS dataset, which operates on a scoring system.  To augment the model's comprehension of pure scores, we incorporated a six-segment evaluation component specific to the Japanese language into the prompt template.  Consequently, this adaptation results in a template that exhibits slight differences from the ones designed for the prior two datasets.

\begin{figure*}[!t]
\centering
\includegraphics[width=450 pt]{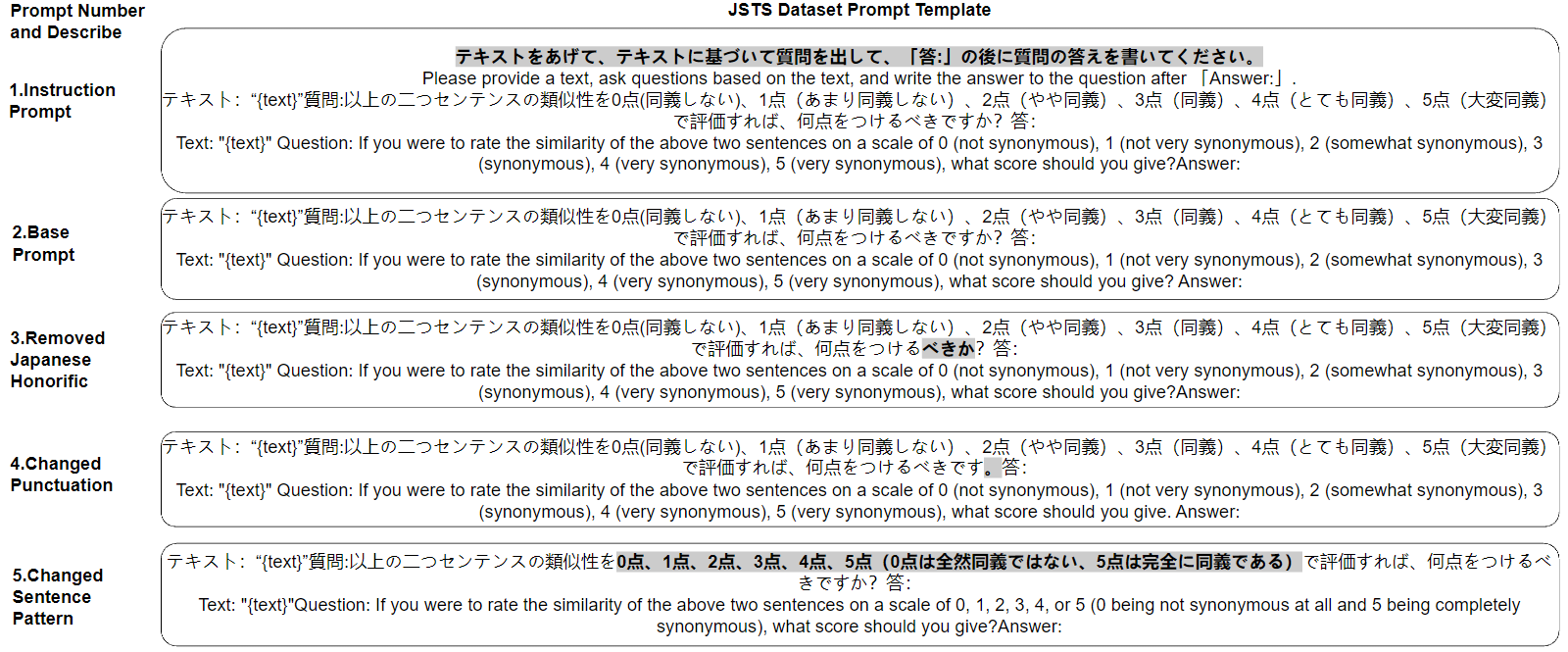}
\caption{\label{9appendixfigure3}The designed prompt template of JSTS dataset.}

\end{figure*}

\section{Generated label extraction algorithm}
\label{sec:algorithm}

As shown in algorithm 1, we design a simple algorithm to extract generated labels and calculate accuracy. First, the total number of samples extracted from each data set is 1000. Then identify and extract the label word from the answer. accuracy is then calculated based on the extracted lable and the actual label.

\begin{algorithm}
\caption{Generated Label Extraction and Accuracy Calculation}
\begin{algorithmic}[1]
\REQUIRE generated\_text\_list, true\_label\_list
\STATE total = 1000
\STATE correct\_count = 0
\FOR{generated\_text in generated\_text\_list}
    \STATE remaining\_text = split(generated\_text, 'Answer:')[-1].strip()
    \IF{'positive' in remaining\_text}
        \STATE label = 'positive'
    \ELSIF{'negative' in remaining\_text}
        \STATE label = 'negative'
    \ELSE
        \STATE label = 'None'
    \ENDIF
    \IF{label == true\_label\_list[i]}
        \STATE correct\_count += 1
    \ENDIF
\ENDFOR
\STATE accuracy = correct\_count / total
\RETURN accuracy
\end{algorithmic}
\end{algorithm}

% This is a section in the appendix.

\end{document}